\documentclass[a4paper,fleqn]{cas-sc}

\usepackage{enumitem}
\usepackage[numbers,sort&compress]{natbib}
\usepackage{amsmath,amssymb}
\usepackage{booktabs,array}
\usepackage{graphicx}
\usepackage{lineno}

\hypersetup{
  colorlinks=false,
  pdfborder={0 0 1},
  citebordercolor={0.45 0.80 0.50},
  linkbordercolor={1.00 0.55 0.55},
  urlbordercolor={1 1 1},
  filebordercolor={1 1 1}
}

\begin{document}
\let\WriteBookmarks\relax
\def\floatpagepagefraction{1}
\def\textpagefraction{.001}

\shorttitle{CoRe-SAM3}
\shortauthors{S. Liu et al.}

\title [mode = title]{CoRe-SAM3: Conditional Semantic--Visual Reconciliation for SAM3 Crack Segmentation}

\author[1]{Shipeng Liu}
\ead{lsp@xauat.edu.cn}

\author[2]{Liang Zhao}
\cormark[1]
\ead{zhaoliang@xauat.edu.cn}

\author[3]{Dengfeng Chen}
\ead{chdengf@xauat.edu.cn}

\cortext[cor1]{Corresponding author}

\affiliation[1]{
  organization={School of Mechanical and Electrical Engineering, Xi'an University of Architecture and Technology},
  city={Xi'an},
  postcode={710055},
  state={Shaanxi},
  country={China}}

\affiliation[2]{
  organization={School of Artificial Intelligence and Robotics, Xi'an University of Architecture and Technology},
  city={Xi'an},
  postcode={710055},
  state={Shaanxi},
  country={China}}

\affiliation[3]{
  organization={School of Building Equipment Science and Engineering, Xi'an University of Architecture and Technology},
  city={Xi'an},
  postcode={710055},
  state={Shaanxi},
  country={China}}

\begin{abstract}
Crack segmentation requires a model to recognize target semantics while accurately recovering thin, low-contrast, and topologically continuous local structures. Although SAM3 provides strong open-concept segmentation, its direct application to the crack domain still misses weak cracks, activates crack-like background regions, and produces local boundary errors. We first diagnose the functional differences between the internal prompt-conditioned semantic representation and native visual representation of SAM3 on five crack datasets. The results show that the semantic representation already carries most task information for crack prediction, whereas the utility of the visual representation depends on the current semantic state. Directly combining the two representations does not yield consistent gains. Based on this finding, we propose Conditional Semantic--Visual Reconciliation, termed CoRe. CoRe retains semantic prediction as the primary decision path, applies lightweight semantic calibration to adjust the target-domain decision mapping, and uses spatially aligned native visual evidence to generate a zero-initialized, bounded, and regularized conditional residual that selectively corrects existing predictions. Across five domains, CoRe-SAM3 improves the average Crack IoU from 62.34\% to 70.47\% and clDice from 81.98\% to 89.24\%, while introducing only 18.914 K trainable parameters. Prediction-transition analysis further shows that CoRe corrects an average of 34.38\% of native errors, with a damage rate of only 0.23\% on pixels correctly classified by native SAM3. These results demonstrate that constrained prediction correction based on the functional differences between internal representations provides an effective and parameter-efficient target-domain adaptation strategy for vision foundation models with strong task-specific semantic priors.
\end{abstract}

\begin{keywords}
Crack segmentation \sep Vision foundation model \sep SAM3 \sep Parameter-efficient adaptation \sep Semantic--visual reconciliation
\end{keywords}

\maketitle

\section{Introduction}\label{sec:introduction}

Crack segmentation is a central vision task in intelligent infrastructure inspection. Its objective extends beyond detecting the presence of cracks to recovering their slender geometry, local width, and topological continuity as accurately as possible. Compared with common natural objects, cracks often exhibit low contrast, large scale variation, elongated structures, and a small foreground ratio. Meanwhile, joints, scratches, shadow boundaries, and repetitive textures can form local visual patterns similar to real cracks. The challenge therefore lies not only in target recognition but also in making accurate spatial decisions among complex local cues. A model must recover weak or discontinuous cracks, suppress background structures with similar appearances, and avoid expanding regions that are already predicted correctly. Existing crack-specific models address this problem by enhancing fine-grained representation through convolution--Transformer collaboration \cite{ref1}, boundary modeling \cite{ref3}, and structure-aware representations \cite{ref4}. These methods generally rely on task-specific architectures and relearn discriminative representations from crack data, with the fundamental goal of acquiring crack recognition and spatial recovery capabilities from supervised data.

Vision foundation models change this starting point. From SAM \cite{ref21} and SAM2 \cite{ref22} to SAM3 \cite{ref23}, which further supports concept-level prompts, large-scale pretraining provides strong general visual and semantic knowledge before downstream training. Given the text prompt \texttt{crack}, SAM3 already produces a semantic response clearly associated with cracks. Downstream adaptation therefore need not learn a complete crack detector from scratch and can instead build on the existing semantic decision. The remaining errors are mainly local spatial deviations in the semantic judgment, including insufficient responses to low-contrast or thin cracks, false activation of crack-like regions such as joints and textures, and response expansion or width errors in partially detected regions. For SAM3, which already has strong pretrained semantic capability, crack adaptation consequently shifts toward preserving reliable semantic decisions while using finer-grained image evidence to correct a limited set of critical local errors.

A direct way to implement such local correction is to introduce an additional visual representation beyond the semantic prediction of SAM3. For example, an independent visual encoder can extract texture, edge, or multiscale structural information from the input image and combine these features with the SAM3 semantic representation for prediction. Local appearance cues may compensate for insufficient semantic responses to thin cracks and determine whether activations on joints, scratches, and other crack-like regions receive adequate visual support. However, a new visual encoding path increases model dependencies and computational cost and may redundantly encode image information already extracted during the SAM3 forward pass. This raises a more fundamental question: does SAM3 itself already contain sufficient visual evidence for local correction?

The answer lies within the model. SAM3 retains internal representations with different functional properties while producing its final semantic response \cite{ref23}. Prompt-conditioned pixel embeddings are conditioned on text and directly carry semantic evidence related to the crack concept. In parallel, FPN features from the visual backbone preserve local appearance and spatial structure closer to the input image. SAM3 thus provides both a semantic state that expresses how the model currently identifies cracks and native visual evidence that describes the local image content. The question consequently narrows from whether additional visual information is needed to whether the existing visual information can correct semantic prediction and how it should participate in this correction.

To answer this question, we probe the prompt-conditioned semantic representation and native visual representation of SAM3 across five crack domains. The semantic representation exhibits substantially stronger crack-label separability on all five datasets, confirming that it is the primary carrier of task information for SAM3 crack prediction. In comparison, the native visual representation has limited discriminative power both for independent crack prediction and for identifying errors in semantic prediction. More importantly, directly combining the two representations for error prediction performs worse than using the semantic representation alone on four of the five datasets. This observation shifts the focus of SAM3 crack adaptation toward heterogeneous evidence utilization. Once the model has formed a strong semantic belief, the semantic representation should retain responsibility for the main crack decision, while the native visual representation should provide conditional support that determines whether and how the semantic response is modified.

\begin{figure}[pos=t]
\centering
\includegraphics[width=0.98\textwidth]{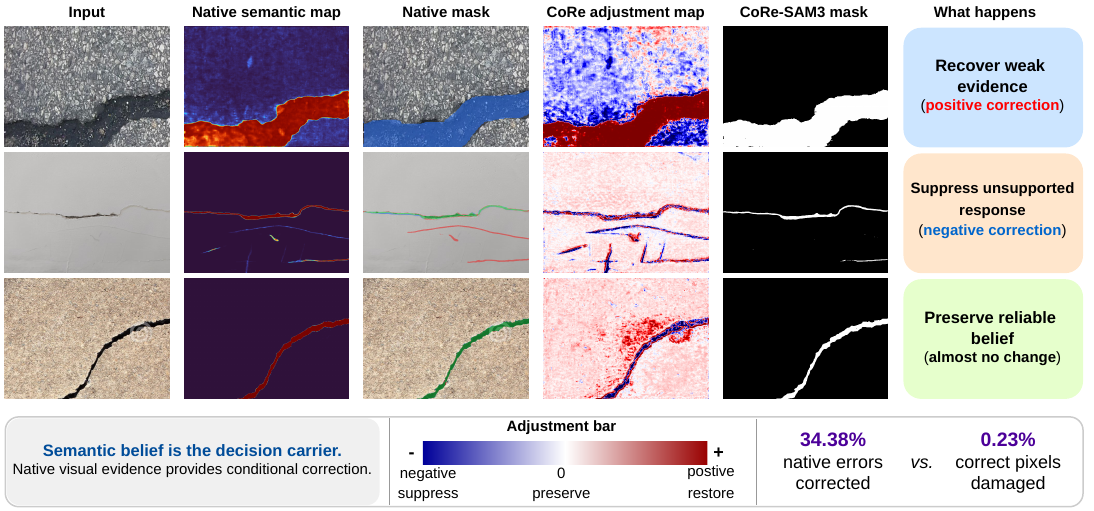}
\caption{Qualitative visualization of CoRe-SAM3 response correction. From left to right, the columns show the input image, native SAM3 semantic map, native mask, signed CoRe adjustment map, and final CoRe-SAM3 mask. In the color-coded mask overlays, green denotes true-positive crack pixels, blue denotes false negatives, and red denotes false positives. The native semantic map represents the continuous prompt-conditioned crack response. In the signed adjustment map, red indicates positive correction, blue indicates negative correction, and near-white regions indicate negligible changes. From top to bottom, the examples illustrate recovery of weak crack evidence, suppression of unsupported crack-like responses, and preservation of reliable predictions. The inset reports a 34.88\% correction rate on native errors and a 0.23\% damage rate on pixels correctly classified by native SAM3.}
\label{fig:1}
\end{figure}

Based on this finding, we propose CoRe, short for Conditional Semantic--Visual Reconciliation, which formulates SAM3 crack adaptation as a conditional semantic--visual reconciliation problem. CoRe retains the existing SAM3 semantic prediction as the primary decision and uses the current semantic state to condition a limited residual correction generated from native visual evidence. The visual evidence therefore controls how the existing semantic decision is adjusted. Fig.~\ref{fig:1} presents the input, native semantic response, native mask, CoRe adjustment, and final mask from left to right. The three rows illustrate weak-crack recovery, suppression of erroneous responses, and preservation of reliable predictions.

CoRe decomposes target-domain adaptation into two levels. It first applies lightweight target-domain calibration to the native semantic projection of SAM3 \cite{ref23}, allowing the existing prompt-conditioned representation to better accommodate the global response distribution and decision bias of the crack domain. It then predicts a local reconciliation residual from the prompt-conditioned semantic feature and the native visual FPN feature at the same spatial scale and applies this residual to the calibrated semantic logit. The residual branch uses zero initialization and bounded updates, so training starts from the original SAM3 semantic belief and organizes the new capacity as an incremental correction to the existing decision. Semantic calibration adjusts the overall response distribution and decision bias of pretrained semantic prediction in the target crack domain, while conditional semantic--visual reconciliation uses existing visual evidence within SAM3 to correct errors in the calibrated decision. Across five domains, CoRe-SAM3 improves the average Crack IoU over native SAM3 by 8.13 percentage points and achieves the best overall performance among the representative recent SAM3 adaptation methods considered in our experiments.

This paper makes three main contributions. First, it introduces a diagnosis-driven perspective on SAM3 adaptation for crack segmentation. Our systematic diagnosis confirms that the prompt-conditioned semantic representation is the core task representation for crack prediction and shows that the utility of native visual evidence depends on its conditional relation to the current semantic state. This finding reframes SAM3 domain adaptation as an internal evidence reconciliation problem and provides a direct basis for adaptation design. Second, we propose CoRe-SAM3, which organizes target-domain learning into semantic decision calibration and conditional residual reconciliation. CoRe-SAM3 retains SAM3 semantic prediction as the primary decision path and conditions a zero-initialized, bounded, lightweight residual update on the current semantic state and native visual evidence, yielding an asymmetric decision mechanism centered on the semantic prediction. Third, we conduct systematic experiments on five crack datasets. Compared with native SAM3, CoRe-SAM3 improves the average Crack IoU from 62.34\% to 70.47\%. Prediction transitions, continuous adjustments, constraint ablations, and qualitative results further validate its selective correction mechanism: the conditional residual applies controlled corrections to existing prediction errors and produces a net positive transition in prediction states.

\section{Related Work}\label{sec:related-work}

\subsection{Crack and Curvilinear Structure Segmentation}\label{sec:crack-segmentation}

Crack segmentation faces persistent challenges arising from sparse foregrounds, large scale variation, low contrast, and fragile topological continuity. Recent specialized networks mainly improve local texture extraction, long-range dependency modeling, and boundary-structure modeling. DTrC-Net \cite{ref1} combines Transformer and convolutional features for complex crack scenes. A boundary-aware convolution--Transformer network \cite{ref3} emphasizes the importance of edge constraints for localizing thin cracks. CrackFormer \cite{ref4} improves pixel-level segmentation through hierarchical representations tailored to crack morphology. CrackSeg9k \cite{ref2} consolidates and benchmarks multiple crack datasets and segmentation frameworks, providing a more systematic basis for cross-scene evaluation.

With the adoption of state-space models in vision, crack segmentation increasingly uses Mamba-style architectures to model long-range dependencies in elongated structures. Vision Mamba-based crack segmentation \cite{ref5} explores unified modeling of concrete, asphalt, and masonry surfaces. Topology-aware Mamba \cite{ref6} explicitly incorporates topology into crack representations, while Mamba Meets Crack Segmentation \cite{ref7} further demonstrates the suitability of state-space modeling for continuous crack structures. More recent methods include SCSegamba \cite{ref8}, which emphasizes lightweight structure-aware modeling; MixerCSeg \cite{ref9}, which improves efficiency through decoupled Mamba attention; and RIFT \cite{ref10}, which revisits efficient crack segmentation through task-aligned structural and directional modeling. LiteCrackSeg \cite{ref11} and CONTI-CrackNet \cite{ref12} improve the accuracy--efficiency trade-off through a lightweight hybrid architecture and continuity-aware state-space modeling, respectively.

Beyond these general architectures, HACNet V2 \cite{ref18} revisits full-resolution crack representations. SECrackSeg \cite{ref19} combines a SAM2 S-Adapter with edge-aware mechanisms for crack scenes, while SGL-Mamba \cite{ref20} jointly models global and local structures. These methods show that crack-specific models commonly recover fine-grained structures by strengthening task-specific representations. CoRe instead examines the functional role of visual evidence already present within a vision foundation model when the model has formed a strong semantic representation of cracks.

\subsection{Promptable Segmentation Foundation Models and Downstream Adaptation}\label{sec:promptable-models}

SAM \cite{ref21} combines large-scale mask pretraining with promptable segmentation and establishes a general prompt-driven segmentation paradigm. SAM2 \cite{ref22} extends this capability to unified image and video segmentation. SAM3 \cite{ref23} further supports concept-prompted segmentation based on text phrases and visual examples and uses a shared visual backbone for detection, segmentation, and tracking. To improve the foundation segmentation capability itself, HQ-SAM \cite{ref24} introduces a high-quality output token for finer masks. EfficientSAM \cite{ref30}, MobileSAM \cite{ref31}, and FastSAM \cite{ref32} reduce the cost of general segmentation models through efficient pretraining, lightweight design, and real-time segmentation, respectively.

Another line of work extends the task interfaces and contextual capabilities of general segmentation models. Semantic-SAM \cite{ref33} combines arbitrary-granularity segmentation with semantic recognition, SEEM \cite{ref34} supports unified segmentation and recognition under multiple prompt types, and SegGPT \cite{ref35} formulates segmentation as a general visual task conditioned on context. In few-shot or training-constrained settings, PerSAM \cite{ref28} personalizes SAM from a single example, while Matcher \cite{ref29} performs one-shot segmentation through all-purpose visual feature matching. These studies demonstrate that prompts, context, and existing visual representations can directly reorganize the downstream capabilities of pretrained segmentation models.

Under substantial domain shifts, existing studies often enhance target-domain representations through adapters or parameter-efficient tuning. SAM-Adapter \cite{ref25} injects lightweight task-specific adapters while freezing the main network. MedSAM \cite{ref26} adapts SAM to medical segmentation using large-scale medical image data. CAT-SAM \cite{ref27} uses conditional tuning for few-shot adaptation. SAM2-Adapter \cite{ref36} extends the adapter paradigm to diverse downstream scenarios, while MCA-SAM \cite{ref37} improves discriminability at both token and sample levels through multiscale contrastive adapter learning. Conv-LoRA \cite{ref38} combines convolutional inductive biases with low-rank tuning to improve SAM adaptation to specialized domains with few trainable parameters.

These methods adapt foundation models to new data distributions by introducing trainable parameters, modifying intermediate features, or strengthening target-domain representations. They generally do not explicitly distinguish the functional roles of different internal representations in the existing prediction. Our diagnosis of SAM3 crack segmentation shows that the challenge is not simply the absence of a stronger target-domain representation. The semantic and native visual representations have different information properties and decision values. CoRe therefore reorganizes their roles in prediction correction according to these functional differences: semantic evidence retains control of the main decision, while visual evidence conditionally corrects locally unreliable responses.

\subsection{Parameter-Efficient Adaptation and Reuse of Internal Evidence}\label{sec:parameter-efficient}

As vision foundation models grow, full fine-tuning incurs substantial training and storage costs and may disrupt general representations acquired during pretraining. Parameter-efficient adaptation has therefore become an important route for transferring foundation models. Its central principle is to freeze as much of the pretrained model as possible while introducing only a small number of learnable parameters to adjust representations or decision mappings for the target task. Visual Prompt Tuning \cite{ref39} inserts learnable prompt tokens into the input sequence to adjust the task response of a frozen vision Transformer. AdaptFormer \cite{ref40} introduces lightweight parallel adapters into Transformer blocks to support downstream adaptation with low parameter overhead. SSF \cite{ref41} shows that feature-wise scaling and shifting of frozen features can also effectively change downstream behavior. Other methods directly constrain the weight-update space. LoRA \cite{ref42} approximates parameter updates with low-rank increments, while FacT \cite{ref43} further compresses the task-specific update space of a vision Transformer through tensor decomposition. Although their parameterizations differ, these methods share the principle of retaining the general capabilities acquired during pretraining and learning only the limited increments required by the target task.

Existing parameter-efficient methods mainly study how to inject these increments through prompts, adapters, scale-and-shift operations, or low-rank weight updates. When a foundation model contains multiple internal representations with substantially different semantic properties, spatial granularity, and task relevance, whether these representations should play equivalent roles in downstream decisions rarely serves as a direct basis for adapter design. This distinction is crucial for SAM3 crack segmentation. The prompt-conditioned semantic representation already contains a strong target-semantic judgment, whereas the native visual representation preserves local appearance and spatial evidence that has not been fully compressed by the final task decision. Ignoring this functional difference and directly concatenating, fusing, or jointly enhancing the two representations does not guarantee more reliable crack predictions.

This paper therefore asks not only how to modify SAM3 with few parameters, but also which existing information should be adjusted and how different internal representations should participate in that adjustment. The representation probing in Section~\ref{sec:preliminary} quantifies the relative ability of semantic and native visual representations to discriminate crack labels and characterize prediction errors, thereby identifying their asymmetric functions during adaptation. Semantic evidence remains the primary decision carrier and undergoes lightweight calibration of the target-domain prediction mapping. Native visual evidence does not act as a second independent prediction source; instead, it supplements local residual correction under the current semantic state. CoRe thus instantiates the parameter-efficient principle of preserving the pretrained model and learning limited increments as a constrained prediction-correction mechanism grounded in the functional differences between internal representations.

\section{Preliminary: Semantic--Visual Reconciliation in SAM3}\label{sec:preliminary}

\subsection{Crack-Prediction Errors in SAM3}\label{sec:sam3-errors}

Given an input image $I$ and text prompt $q=\text{\texttt{crack}}$, SAM3 \cite{ref23} directly produces a pixel-level response associated with crack semantics. This process exploits general visual and semantic knowledge acquired through large-scale pretraining and localizes many crack regions without target-domain training. Its errors in crack segmentation are concentrated in local spatial decisions.

These errors fall into three categories. First, semantic evidence may be insufficient for low-contrast, thin, or locally discontinuous cracks, resulting in missing branches or overlooked fine cracks. Second, crack-like patterns such as joints, scratches, and shadow boundaries may receive high semantic responses and produce false-positive regions. Third, correctly identified crack bodies may still exhibit boundary errors. The main challenge for SAM3 in crack segmentation is therefore to convert its strong semantic recognition of cracks into precise and stable local spatial decisions.

This observation motivates the introduction of additional image evidence into semantic prediction. Weak cracks require fine-grained local appearance cues to determine whether their semantic responses should be strengthened. Crack-like background regions require image texture and local structure to determine whether their current responses receive sufficient visual support. An additional visual encoder can extract local or multiscale features and jointly model them with SAM3 semantic features, but it also increases dependencies and computational cost. SAM3 already encodes the input image before producing the semantic response, making its native visual representation a more direct information source. We therefore first examine how the native visual representation of SAM3 can assist crack-semantic decisions.

\subsection{Semantic and Visual Evidence in SAM3}\label{sec:semantic-visual-evidence}

The final crack response of SAM3 is produced through multiple stages of internal computation, in which the same input forms two representations with different properties. We denote them as
\begin{equation}
S=F_{\mathrm{sem}}(I,q), \qquad V=F_{\mathrm{vis}}(I), \label{eq:1}
\end{equation}
where $S\in\mathbb{R}^{C\times H\times W}$ denotes the prompt-conditioned pixel embeddings produced by the Pixel Decoder, and $V\in\mathbb{R}^{C\times H\times W}$ denotes the high-resolution FPN feature from the visual backbone at the same spatial scale.

Although $S$ and $V$ correspond to the same input image and share the same spatial grid, they follow different information-processing paths. $S$ is conditioned on the text prompt $\text{\texttt{crack}}$ and is organized into a representation directly related to the target semantics. $V$ retains local texture, edges, morphology, and spatial structure from native visual encoding. We therefore interpret $S$ and $V$ as semantic evidence and native visual evidence, respectively.

This internal structure makes $V$ a candidate source of information for identifying crack-prediction errors. To determine the functional relationship between $S$ and $V$, we examine two questions: the ability of $V$ relative to $S$ to predict crack labels, and the ability of $V$ to identify local errors in semantic prediction together with the complementary gain from directly combining the two feature types.

We answer these questions through a unified probing analysis of frozen $S$ and $V$. All probes use the same capacity and training protocol so that performance differences primarily reflect the information properties of the representations.

\subsection{Representation Diagnosis and Design Implications}\label{sec:representation-diagnosis}

\begin{figure}[pos=t]
\centering
\includegraphics[width=0.98\textwidth]{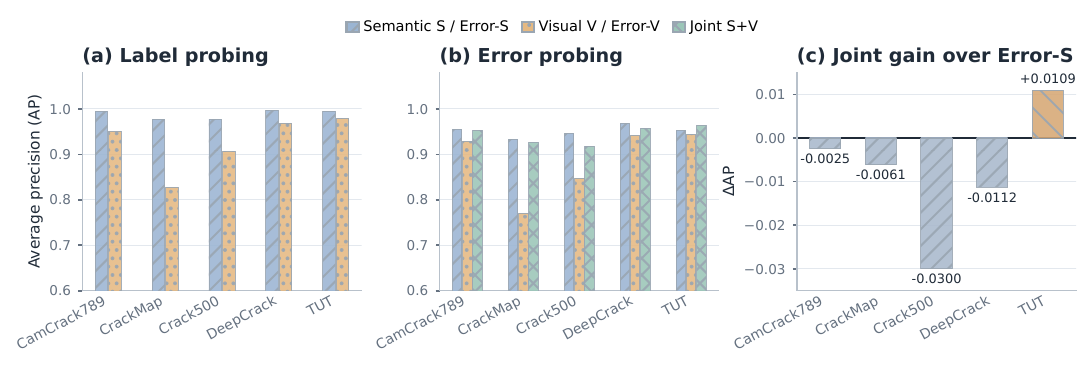}
\caption{Representation diagnostics across five crack datasets: CamCrack789, CrackMap, Crack500, DeepCrack, and TUT. (a) Label probing compares the average precision (AP) of semantic ($S$) and visual ($V$) representations for predicting crack labels. (b) Error probing evaluates error prediction using semantic cues (Error-S), visual cues (Error-V), and their joint representation (Error-S+V). (c) Joint gain over Error-S reports $\Delta\mathrm{AP}=\mathrm{AP}(\text{Error-S+V})-\mathrm{AP}(\text{Error-S})$, where positive and negative values indicate improvement and degradation, respectively. Light-blue diagonal, light-orange dotted, and light-green cross-hatched bars denote semantic or Error-S, visual or Error-V, and joint S+V or Error-S+V, respectively. Panels (a) and (b) use a truncated AP axis with a lower bound of 0.6 for readability.}
\label{fig:2}
\end{figure}

Fig.~\ref{fig:2}(a) first examines the ability of the two representations to directly predict crack labels $Y$. We train two lightweight probes of identical capacity, $S\rightarrow Y$ and $V\rightarrow Y$, and evaluate them on five data domains. $S$ consistently outperforms $V$ on all five datasets. Their mean AP values across the five domains are
\begin{equation}
\operatorname{AP}(S\rightarrow Y)=0.9879, \quad \operatorname{AP}(V\rightarrow Y)=0.9260. \label{eq:2}
\end{equation}

This result shows that the prompt-conditioned representation of SAM3 strongly encodes the semantic information required for crack prediction and serves as the primary task representation in the target domains. The native visual feature is better suited to providing complementary evidence under the current semantic state.

We next examine whether $V$ can act as an additional detector of errors in semantic prediction, as shown in Fig.~\ref{fig:2}(b). We fix the native SAM3 output, construct a pixel-level error label $E$ from its disagreement with the ground truth, and train three probes:
\begin{equation}
S\rightarrow E, \qquad V\rightarrow E, \qquad [S,V]\rightarrow E. \label{eq:3}
\end{equation}

The five-domain results again show that the semantic representation alone has high separability for prediction errors, with $\operatorname{AP}(S\rightarrow E)=0.9509$ and $\operatorname{AP}(V\rightarrow E)=0.8862$. Thus, while encoding crack locations, $S$ also retains substantial information about the reliability of the current prediction. The joint error predictor obtains $\operatorname{AP}([S,V]\rightarrow E)=0.9431$, which is 0.0078 lower than that of the $S\rightarrow E$ probe on average across the five domains. Fig.~\ref{fig:2}(c) reports the per-domain gain of the joint representation over Error-S and shows a small positive gain on only one domain. Simple concatenation therefore fails to establish stable semantic--visual complementarity across domains.

These results yield three diagnostic conclusions. First, the prompt-conditioned representation of SAM3 is the primary information carrier for crack segmentation. Second, the native visual representation is weaker than the semantic representation in both independent crack prediction and error discrimination. Third, the value of visual evidence must be organized through its conditional relationship with the current semantic state. Effective SAM3 adaptation for crack segmentation should therefore reconcile existing visual information within the model by retaining the dominant role of semantic prediction and using native visual evidence to condition how the semantic decision is adjusted in the target domain.

This diagnosis transforms the problem from conventional feature augmentation into evidence reconciliation. Instead of directly learning $[S,V]\rightarrow Z_{\mathrm{new}}$, a more suitable objective retains and calibrates the existing SAM3 semantic prediction before allowing the two forms of internal evidence to determine the required correction:
\begin{equation}
Z_{\mathrm{final}}=Z_{\mathrm{cal}}+\Delta Z_{\mathrm{rec}}(S,V), \label{eq:4}
\end{equation}
where $Z_{\mathrm{cal}}$ denotes the target-domain-calibrated semantic belief and $\Delta Z_{\mathrm{rec}}$ is responsible only for reconciling local decisions supported by semantic and native visual evidence. We consequently focus on using native visual evidence already present within SAM3 to apply conditional and constrained corrections to its semantic judgment of cracks.

Based on this diagnosis, CoRe-SAM3 retains the semantic representation as the primary decision carrier, uses the native visual representation as conditional support, and applies lightweight residual reconciliation to correct target-domain prediction errors. The next section details this process.

\section{Method}\label{sec:method}

\begin{figure}[pos=t]
\centering
\includegraphics[width=0.98\textwidth]{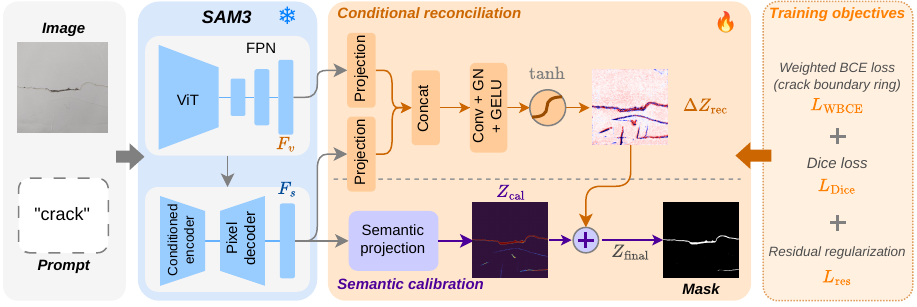}
\caption{Overview of CoRe-SAM3. CoRe adapts SAM3 for crack segmentation by preserving the prompt-conditioned semantic prediction as the primary decision path and exploiting native visual evidence for conditional residual correction. The frozen SAM3 backbone produces semantic features $F_s$ and native visual features $F_v$. Semantic calibration first adjusts the target-domain decision mapping, while the conditional reconciliation branch predicts a bounded residual $\Delta Z_{\mathrm{rec}}$ from joint semantic--visual evidence. The final prediction conservatively updates the calibrated semantic belief with this residual. Boundary-aware weighted BCE loss, Dice loss, and residual regularization jointly balance effective correction against the magnitude of prediction changes.}
\label{fig:3}
\end{figure}

The representation diagnosis in the previous section shows that SAM3 adaptation for crack segmentation must coordinate the roles of existing heterogeneous evidence in target-domain decisions. The prompt-conditioned semantic representation provides strong crack-label separability and retains substantial information for error discrimination. The native visual representation is weaker in independent prediction and error discrimination, and direct combination does not produce stable complementarity. Based on this capability relationship, CoRe retains the semantic representation as the primary decision source, lightly calibrates its target-domain readout, and uses native visual evidence to condition a limited correction of the remaining local errors. CoRe formulates SAM3 target-domain adaptation as
\begin{equation}
Z_{\mathrm{final}}=Z_{\mathrm{cal}}+\Delta Z_{\mathrm{rec}}, \label{eq:5}
\end{equation}
where $Z_{\mathrm{cal}}$ denotes the target-domain-calibrated output produced from the prompt-conditioned semantic representation, and $\Delta Z_{\mathrm{rec}}$ denotes a constrained local correction jointly determined by semantic and native visual evidence. Throughout this process, the image backbone, text or prompt encoder, conditioned encoder, and Pixel Decoder of SAM3 \cite{ref23} remain frozen. Only the native semantic projection and the lightweight reconciliation branch are updated. As shown in Fig.~\ref{fig:3}, the model sequentially performs frozen feature extraction, semantic calibration, conditional reconciliation, and loss supervision, then adds the bounded residual to the calibrated semantic logit to obtain the final mask.

\subsection{Diagnosis-Driven Adaptation Framework}\label{sec:framework}

Given an input image $I$ and a fixed text prompt $q=\texttt{crack}$, frozen SAM3 produces two spatially aligned internal representations:
\begin{equation}
F_s=F_{\mathrm{sem}}(I,q) \in \mathbb{R}^{C\times H\times W}, \quad F_v=F_{\mathrm{vis}}(I) \in \mathbb{R}^{C\times H\times W}, \label{eq:6}
\end{equation}
where $F_s$ denotes the prompt-conditioned pixel embeddings from the Pixel Decoder and $F_v$ denotes the high-resolution FPN feature from the visual backbone at the same spatial scale. Although both representations originate from the same input and share the same spatial dimensions, they assume different decision roles in CoRe. $F_s$ is conditioned on the text prompt and exhibits substantially stronger label separability in the diagnosis of Section~\ref{sec:preliminary}, so CoRe directly retains it as the basis of the final decision. $F_v$ preserves more native visual information and serves as conditional evidence under the current semantic state to determine whether the existing decision requires further adjustment. CoRe learns $F_s\rightarrow Z_{\mathrm{cal}}$ and $(F_s,F_v)\rightarrow\Delta Z_{\mathrm{rec}}$, yielding
\begin{equation}
Z_{\mathrm{final}}=Z_{\mathrm{cal}}+\Delta Z_{\mathrm{rec}}. \label{eq:7}
\end{equation}

The asymmetry in CoRe thus arises from the distinct functions of the two representations in the final prediction path. The semantic branch directly determines the calibrated primary prediction, whereas the semantic--visual branch incrementally modifies this prediction through a constrained residual.

\subsection{Target-Domain Semantic Decision Calibration}\label{sec:semantic-calibration}

The label probing in the previous section shows that the prompt-conditioned representation $F_s$ already provides strong crack-label separability across the five crack domains. A direct adaptation strategy is therefore to adjust how the target-domain decision is read from the existing $F_s$.

The native semantic head of SAM3 is a linear spatial projection applied to $F_s$ by
\begin{equation}
Z_{\mathrm{sam}} = W_s*F_s+b_s, \label{eq:8}
\end{equation}
where $*$ denotes a $1\times1$ convolution, and $W_s$ and $b_s$ are the semantic projection parameters from the SAM3 checkpoint.

Although $F_s$ already contains strong crack-semantic information, crack domains vary substantially in imaging conditions, crack scales, background textures, and class distributions. CoRe freezes the generation of $F_s$ and updates the native semantic projection under target-domain supervision:
\begin{equation}
Z_{\mathrm{cal}} = \widetilde W_s*F_s+\widetilde b_s, \label{eq:9}
\end{equation}
where $\widetilde W_s$ and $\widetilde b_s$ are initialized from the original SAM3 parameters.

This adaptation adjusts the decision mapping from the frozen representation space to the crack-logit space, allowing CoRe to calibrate how it reads the existing crack semantics. The decomposition also reduces the burden on the subsequent reconciliation branch. If the native semantic projection remains fully frozen, the new branch must correct both the basic decision mismatch in the target domain and spatially dependent errors involving low-contrast cracks, crack-like backgrounds, and local boundaries. Semantic decision retargeting allows the residual branch to focus on local decision discrepancies that remain after calibration.

\subsection{Bounded Conditional Semantic--Visual Reconciliation}\label{sec:bounded-reconciliation}

Two types of local errors may remain after target-domain semantic calibration. A true crack may have some semantic evidence but an insufficient response for a correct foreground decision. Conversely, a joint, shadow boundary, or repetitive texture may receive a strong crack response even though its local visual support is inconsistent with a real crack. CoRe conditionally corrects these remaining decisions using the native visual representation already present within SAM3.

CoRe uses $F_s$ to describe the current semantic state and lets $F_v$ regulate the direction and magnitude of a limited correction under this state, producing a local decision update to the existing semantic belief. Because the two features originate from different information paths, they are first projected independently into a shared low-dimensional space by
\begin{equation}
\hat F_s=\phi_s(F_s), \qquad \hat F_v=\phi_v(F_v), \label{eq:10}
\end{equation}
where
\begin{equation}
\phi(\cdot)=\operatorname{GELU}\left(\operatorname{GN}\left(\operatorname{Conv}_{1\times1}(\cdot)\right)\right). \label{eq:11}
\end{equation}

Both projection paths reduce the channel dimension from $256$ to $32$. The two forms of evidence are then concatenated along the channel dimension:
\begin{equation}
F_c=[\hat F_s;\hat F_v] \in \mathbb{R}^{64\times H\times W}, \label{eq:12}
\end{equation}
and a lightweight $1\times1$ projection produces the residual representation:
\begin{equation}
F_r=\operatorname{GELU}\left(\operatorname{GN}\left(W_f*F_c+b_f\right)\right), \label{eq:13}
\end{equation}
where $W_f$ maps the $64$-dimensional joint representation back to $32$ dimensions. These simple independent projections provide the residual estimator with the current semantic state and the corresponding native appearance evidence under a low-capacity design. The residual-update form further constrains the decision authority of visual evidence. CoRe defines the reconciliation residual as
\begin{equation}
\Delta Z_{\mathrm{rec}} = s \tanh \left( W_r*F_r+b_r \right), \label{eq:14}
\end{equation}
where $s$ is the maximum residual scale. Therefore, $\Delta Z_{\mathrm{rec}}(x) \in [-s,s]$ for every pixel $x$, and the final prediction is
\begin{equation}
Z_{\mathrm{final}}(x)=Z_{\mathrm{cal}}(x) + \Delta Z_{\mathrm{rec}}(x). \label{eq:15}
\end{equation}

This bounded update directly follows the representation diagnosis in Section~\ref{sec:preliminary}. Five-domain probing shows that the gain from directly combining visual representations varies markedly across domains, so CoRe explicitly limits their maximum effect on the final logit. Target-domain supervision then drives the reconciliation branch to learn positive or negative corrections for the corresponding semantic--visual configuration.

The residual output parameters $W_r$ and $b_r$ are zero-initialized, so $\Delta Z_{\mathrm{rec}}=0$ at the beginning of training and $Z_{\mathrm{final}}=Z_{\mathrm{cal}}$. Zero initialization preserves semantic prediction at the starting point and lets the new branch gradually learn the required deviation from an identity update. Together with bounded $\tanh$, CoRe constructs reconciliation as a conservative belief update in which pretrained semantic evidence forms the prediction basis and native visual evidence contributes a limited correction under target-domain supervision.

This design also explains why the residual branch retains both $F_s$ and $F_v$. Error probing shows that $F_s$ already contains substantial information about the reliability of the current prediction. Local correction must therefore be conditioned on the semantic state described by $F_s$ and cannot be reduced to
\begin{equation}
\Delta Z=g(F_v). \label{eq:16}
\end{equation}

The model first uses $F_s$ to determine the current semantic state and then uses local visual evidence from $F_v$ to decide whether this state should be strengthened, suppressed, or left unchanged.

\subsection{Optimization Objective and Conservative Residual Regularization}\label{sec:objective}

CoRe uses standard segmentation supervision during training. To address the thin geometry of cracks and their tendency to exhibit boundary expansion, it assigns greater weight to background errors near cracks and applies residual regularization to encourage necessary corrections of small magnitude.

Given a binary crack annotation $Y$, we first define the outer neighborhood of cracks through morphological dilation by
\begin{equation}
\Omega_{\mathrm{ring}} = \operatorname{Dilate}(Y,r)-Y, \label{eq:17}
\end{equation}
and define the weight of each pixel in this outer neighborhood as
\begin{equation}
w(x)=1+ \omega \mathbb I \left[ x\in\Omega_{\mathrm{ring}} \right], \label{eq:18}
\end{equation}
so background pixels immediately adjacent to real cracks receive higher weights during optimization. This weighting increases the sensitivity of pixel classification to local foreground expansion, nearby adhesion, and responses to crack-like backgrounds. Let
\begin{equation}
P(x)=\sigma \left(Z_{\mathrm{final}}(x)\right), \label{eq:19}
\end{equation}
then the weighted binary cross-entropy loss is
\begin{equation}
\mathcal L_{\mathrm{WBCE}} = - \frac{1}{|\Omega|}\sum_{x\in\Omega} w(x) \left[Y(x)\log P(x) + (1-Y(x)) \log(1-P(x)) \right]. \label{eq:20}
\end{equation}

Because crack foreground occupies a small fraction of the image, we also use Dice loss to constrain overall foreground overlap:
\begin{equation}
\mathcal L_{\mathrm{Dice}} = 1- \frac{2\sum_x P(x)Y(x)+\epsilon}{\sum_x P(x)+\sum_xY(x)+\epsilon}. \label{eq:21}
\end{equation}

In addition to segmentation supervision, CoRe directly constrains the magnitude of the reconciliation residual:
\begin{equation}
\mathcal L_{\mathrm{res}} = \frac{1}{|\Omega|} \sum_{x\in\Omega} \Delta Z_{\mathrm{rec}}(x)^2. \label{eq:22}
\end{equation}

This regularizer serves the same purpose as the bounded residual in Section~\ref{sec:bounded-reconciliation}. The $\tanh$ function structurally limits the maximum magnitude of a single correction, while $\mathcal L_{\mathrm{res}}$ favors smaller updates during optimization. Under the segmentation objective, the model therefore modifies only the calibrated semantic predictions that require correction. The final CoRe-SAM3 training objective is
\begin{equation}
\mathcal L = \mathcal L_{\mathrm{WBCE}} + \lambda_d \mathcal L_{\mathrm{Dice}} + \lambda_r \mathcal L_{\mathrm{res}}, \label{eq:23}
\end{equation}
where $\lambda_d$ and $\lambda_r$ are hyperparameters.

Throughout optimization, the image backbone, text or prompt encoder, conditioned encoder, and Pixel Decoder of SAM3 remain frozen. Target-domain learning in CoRe occurs at two points. It first recalibrates the mapping from the frozen prompt-conditioned representation to the crack-decision space, then learns a low-capacity, bounded, and zero-initialized conditional reconciliation residual. The adaptation process prioritizes semantic knowledge already present in SAM3 and applies only the corrections required for the remaining local decision errors in the target domain.

\section{Experiments}\label{sec:experiments}

We evaluate CoRe-SAM3 from five perspectives: cross-domain segmentation performance, validation of key design choices, prediction-correction behavior, qualitative results, and computational efficiency. Section~\ref{sec:preliminary} diagnoses the information properties of different internal representations in SAM3. This section further examines how these diagnostic findings translate into an effective target-domain adaptation strategy. The experiments address three questions: how effectively semantic calibration and conditional semantic--visual reconciliation adapt across crack domains; how retaining the semantic prior, incorporating visual evidence, and constraining the residual contribute to performance; and whether the direction of CoRe adjustments aligns with error types while preserving existing correct decisions.

\subsection{Experimental Setup}\label{sec:setup}

\subsubsection{Datasets}\label{sec:datasets}

We conduct experiments on five public crack-segmentation datasets: CamCrack789 \cite{ref13}, Crack500 \cite{ref16}, CrackMap \cite{ref14}, DeepCrack \cite{ref15}, and TUT \cite{ref17}. These datasets differ substantially in image resolution, background texture, crack scale, and morphological distribution, enabling evaluation of adaptation across diverse crack domains. All compared methods use the same training, validation, and test splits for each dataset. Table~\ref{tab:1} reports the number of samples and original image resolutions.

\begin{table}[pos=t]
\centering
\begin{tabular}{lcccc}
\toprule
\textbf{Dataset} & \textbf{Train} & \textbf{Validation} & \textbf{Test} & \textbf{Image resolution} \\
\midrule
CamCrack789 & 553 & 79 & 157 & $640\times480$ \\
Crack500 & 2357 & 336 & 675 & $(360\sim648)\times(360\sim640)$ \\
CrackMap & 84 & 12 & 24 & $256\times256$ \\
DeepCrack & 368 & 53 & 106 & $(384\sim544)\times(384\sim544)$ \\
TUT & 987 & 139 & 282 & $640\times640$ \\
\bottomrule
\end{tabular}%
\caption{Dataset statistics and original image resolutions for the five crack-segmentation benchmarks. The numbers of images in the training, validation, and test subsets follow the dataset splits used in all experiments.}
\label{tab:1}
\end{table}

Table~\ref{tab:1} shows substantial variation in both sample size and original image resolution across the five datasets. SAM3-based methods uniformly use an input resolution of $1008\times1008$. Table~\ref{tab:2} also includes several dedicated crack-segmentation networks evaluated at $512\times512$ as task-performance references, while controlled comparisons focus on adaptation methods using the same SAM3 setting.

\subsubsection{Implementation Details}\label{sec:implementation}

All SAM3-based experiments use the same pretrained checkpoint and the fixed text prompt \texttt{crack}. CoRe-SAM3 freezes the image backbone, text or prompt encoder, conditioned encoder, and Pixel Decoder, and optimizes only the native semantic projection of SAM3 and the newly introduced reconciliation branch.

The prompt-conditioned feature and native visual FPN feature both have dimensions of $256\times288\times288$. Independent projections reduce each representation to 32 channels before conditional residual prediction. The maximum residual scale is $s=0.5$. The morphological radius and additional weight for the outer crack region are $r=3$ and $\omega=4$, respectively. The weights of Dice loss and residual regularization are $\lambda_d=0.25$ and $\lambda_r=0.05$, respectively.

Training uses the AdamW optimizer with an initial learning rate of $1\times10^{-3}$ and a weight decay of $1\times10^{-4}$. The batch size is 1. We train for at most 60 epochs with early stopping at a patience of 10 and select the model according to Crack IoU on each domain-specific validation set. All results are averaged over three fixed random seeds. The training platform uses Ubuntu 22.04.5 LTS, an Intel Core i9-11900K CPU, 32 GB RAM, and an NVIDIA RTX A2000 GPU with 12 GB memory.

\subsubsection{Evaluation Metrics}\label{sec:metrics}

Crack foreground usually occupies only a small image region, so overall pixel accuracy is easily dominated by background pixels. We use Crack IoU as the primary metric and also report mean IoU (mIoU), F1-score, Precision, and Recall. Because cracks are typical curvilinear topological structures, we further use clDice \cite{ref44} to evaluate the preservation of connectivity and centerline structure. We also define the predicted area ratio (PAR) to measure the global relationship between predicted foreground area and ground-truth crack area:
\begin{equation}
\mathrm{PAR}=|\widehat Y| / |Y|. \label{eq:24}
\end{equation}

A PAR value close to 1 indicates that predicted and annotated foreground areas are similar overall, whereas values above and below 1 indicate global expansion and contraction, respectively. We interpret PAR jointly with Precision, Recall, clDice, and the subsequent prediction-transition analysis to distinguish correct recovery from erroneous expansion.

\subsection{Overall Segmentation Performance}\label{sec:overall-results}

\subsubsection{Performance across Five Crack Domains}\label{sec:five-domain-results}

Table~\ref{tab:2} compares Crack IoU for CoRe-SAM3, native SAM3, existing SAM3 adaptation methods, and dedicated crack-segmentation networks on five crack datasets. It also reports the number of parameters updated during adaptation or training. The comparison assesses whether each method consistently improves SAM3 segmentation across heterogeneous crack domains and how many task-specific parameters are required to obtain this improvement.

\begin{table}[pos=t]
\centering
\resizebox{\textwidth}{!}{%
\begin{tabular}{lccccccc}
\toprule
\textbf{Method} & \textbf{Trainable Params} & \textbf{CamCrack789} & \textbf{Crack500} & \textbf{CrackMap} & \textbf{DeepCrack} & \textbf{TUT} & \textbf{Avg.} \\
\midrule
\multicolumn{8}{l}{\textit{\textbf{Crack specialist}}} \\
\addlinespace[1pt]
~~SCSegamba \cite{ref8} & 2.80 M & 66.65 & 58.43 & 62.20 & 81.22 & 70.23 & 67.75 \\
~~MixerCSeg \cite{ref9} & 2.54 M & 69.20 & 59.37 & 63.76 & 83.24 & 64.62 & 68.04 \\
~~RIFT-T \cite{ref10} & 0.47 M & 72.20 & 61.42 & 64.09 & 83.74 & 68.78 & 70.04 \\ \midrule
\multicolumn{8}{l}{\textit{\textbf{SAM3-based adaptation}}} \\
\addlinespace[1pt]
~~HAT-SAM3 \cite{ref45} & 85.74 M & 62.74 & 57.44 & 48.95 & 76.75 & 64.15 & 62.01 \\
~~SAM3 \cite{ref23} & -- & 70.48 & 41.33 & 55.61 & 80.25 & 64.03 & 62.34 \\
~~SAM3-adapter \cite{ref46} & 4.18 M & 65.70 & 57.65 & 48.52 & 76.85 & 64.73 & 62.69 \\
~~SAM3-UNet \cite{ref47} & 2.77 M & 69.58 & \textbf{61.21} & 63.49 & 80.06 & 68.69 & 68.61 \\
~~CoRe-SAM3 (ours) & \textbf{18.91 K} & \textbf{74.71} & 57.28 & \textbf{65.78} & \textbf{83.86} & \textbf{70.70} & \textbf{70.47} \\
\bottomrule
\end{tabular}%
}
\caption{Overall segmentation performance across five crack datasets. The table reports Crack IoU (\%) for CoRe-SAM3, native SAM3, SAM3-based adaptation methods, and dedicated crack-segmentation networks. Results of dedicated crack-segmentation networks serve as task-level performance references because their architectures, pretraining schemes, and input resolutions differ from those of SAM3-based methods.}
\label{tab:2}
\end{table}

CoRe-SAM3 achieves higher Crack IoU than training-free native SAM3 on all five datasets. The five-domain average increases from 62.34\% to 70.47\%, an absolute gain of 8.13 percentage points. The improvements on CamCrack789, Crack500, CrackMap, DeepCrack, and TUT are 4.23, 15.95, 10.17, 3.61, and 6.67 percentage points, respectively. These consistent gains show that CoRe is not restricted to a particular crack distribution and reliably corrects the original SAM3 predictions across domains with different background complexity, crack scales, and imaging conditions.

CoRe-SAM3 also exhibits a clear parameter-efficiency advantage among SAM3-based adaptation methods. It introduces only 18.91 K trainable parameters, substantially fewer than the 85.74 M of HAT-SAM3 \cite{ref45}, 4.18 M of SAM3-adapter \cite{ref46}, and 2.77 M of SAM3-UNet \cite{ref47}. SAM3-UNet and SAM3-adapter therefore use approximately 146 and 221 times as many trainable parameters as CoRe-SAM3, respectively, while HAT-SAM3 uses more than 4500 times as many. Despite this small adaptation space, CoRe-SAM3 achieves the best result among SAM3-based methods on CamCrack789, CrackMap, DeepCrack, and TUT, and its mean Crack IoU of 70.47\% exceeds that of the larger SAM3-UNet by 1.86 percentage points. Improving SAM3 adaptation for crack segmentation therefore does not necessarily require large-scale parameter updates or a high-capacity decoder. Effective use of existing internal representations to correct the original prediction is more important.

Crack500 is the only dataset on which CoRe-SAM3 does not achieve the best result among SAM3-based methods. Although CoRe-SAM3 substantially improves native SAM3 from 41.33\% to 57.28\%, SAM3-UNet \cite{ref47} reaches 61.21\%. When the target domain requires substantial spatial reconstruction relative to the original response of the foundation model, a higher-capacity decoder may retain an advantage. CoRe does not aim to replace such architectures by relearning a complete segmentation mapping. It corrects the existing prediction with a small task-specific parameter budget while preserving pretrained SAM3 capabilities. The Crack500 result therefore delineates the capability boundary between lightweight constrained correction and high-capacity target-domain reconstruction.

Dedicated crack networks provide an additional task-level performance reference. CoRe-SAM3 achieves a five-domain mean Crack IoU of 70.47\%, slightly above the 70.04\% of RIFT-T \cite{ref10}, and obtains higher values on CamCrack789, CrackMap, DeepCrack, and TUT. RIFT-T is itself a highly lightweight crack-specific network with 0.47 M trainable parameters, approximately 25 times the number newly introduced by CoRe-SAM3. The two model classes do not share identical training or computational conditions because dedicated networks and SAM3-based methods differ fundamentally in foundation-model scale, pretraining, and inference path. This comparison instead shows that targeted correction with very few task-specific parameters can achieve competitive crack-segmentation performance when built on an existing foundation-model representation.

\subsubsection{Multi-Metric Evaluation}\label{sec:multi-metric}

To further characterize the performance changes, Table~\ref{tab:3} reports the five-domain macro-average results of native SAM3 and CoRe-SAM3 under multiple metrics.

\begin{table}[pos=t]
\centering
\begin{tabular}{lccccccc}
\toprule
\textbf{Method} & \textbf{Crack IoU} $\uparrow$ & \textbf{mIoU} $\uparrow$ & \textbf{F1} $\uparrow$ & \textbf{Precision} $\uparrow$ & \textbf{Recall} $\uparrow$ & \textbf{PAR} & \textbf{clDice} $\uparrow$ \\
\midrule
SAM3 & 62.34 & 80.33 & 73.14 & 74.70 & 74.16 & 0.939 & 81.98 \\
CoRe-SAM3 (ours) & 70.47 & 84.53 & 81.58 & 79.94 & 85.60 & 1.124 & 89.24 \\
$\Delta$ & \textbf{+8.13} & \textbf{+4.20} & \textbf{+8.44} & \textbf{+5.24} & \textbf{+11.44} & -- & \textbf{+7.26} \\
\bottomrule
\end{tabular}%
\caption{Multi-metric comparison between native SAM3 and CoRe-SAM3 across five crack datasets. Values are macro-averaged over the five datasets. Improvements are absolute differences between CoRe-SAM3 and native SAM3. PAR denotes the ratio of predicted foreground area to ground-truth foreground area.}
\label{tab:3}
\end{table}

Table~\ref{tab:3} shows that CoRe-SAM3 substantially improves mIoU, F1, Precision, Recall, and clDice in addition to Crack IoU. Recall increases from 74.16\% to 85.60\%, while Precision improves by 5.24 percentage points. Target-domain adaptation therefore recovers weak crack regions while improving foreground discrimination.

CoRe-SAM3 improves clDice from 81.98\% to 89.24\%, indicating that the added foreground responses generally correspond to more complete crack topology. PAR increases from 0.939 to 1.124, showing that recovery of missing cracks changes the overall predicted area from slight underestimation to moderate foreground expansion. Taken together, the metrics indicate both missed-crack recovery and suppression of erroneous responses. Section~\ref{sec:correction-analysis} further validates this behavior through pixel-state transitions and the direction of prediction adjustments.

\subsection{Validation of the Reconciliation Design}\label{sec:design-validation}

The diagnosis in Section~\ref{sec:preliminary} shows that the prompt-conditioned semantic representation is the primary task-information carrier, whereas gains from directly concatenating the native visual representation are inconsistent across domains. We therefore validate target-domain semantic calibration, the form in which visual evidence participates, and the constraints on residual correction in the same order as the CoRe design process.

\subsubsection{From Semantic Calibration to Conditional Reconciliation}\label{sec:incremental-design}

Table~\ref{tab:4} progressively introduces semantic calibration, semantic residual correction, and native visual evidence on CamCrack789 and CrackMap.

\begin{table}[pos=t]
\centering
\resizebox{\textwidth}{!}{%
\begin{tabular}{ccccccccc}
\toprule
\multirow{2}{*}{\textbf{Semantic calibration}}
& \multirow{2}{*}{\textbf{Semantic residual}}
& \multirow{2}{*}{\textbf{Visual evidence}}
& \multicolumn{3}{c}{\textbf{CamCrack789}}
& \multicolumn{3}{c}{\textbf{CrackMap}} \\
\cmidrule(lr){4-6}\cmidrule(lr){7-9}
& & & \textbf{Crack IoU} & \textbf{PAR} & \textbf{clDice} & \textbf{Crack IoU} & \textbf{PAR} & \textbf{clDice} \\
\midrule
 &  &  & 70.48 & 1.129 & 93.50 & 55.61 & 0.843 & 77.97 \\
$\checkmark$ &  &  & 72.98 & 1.043 & 93.77 & 59.16 & 1.247 & 84.96 \\
$\checkmark$ & $\checkmark$ &  & 73.53 & 1.052 & 93.74 & 63.30 & 1.257 & 85.14 \\
$\checkmark$ & $\checkmark$ & $\checkmark$ & 74.71 & 1.034 & 94.69 & 65.78 & 1.223 & 89.09 \\
\bottomrule
\end{tabular}%
}
\caption{Incremental evaluation of semantic calibration and semantic--visual reconciliation. Starting from native SAM3, semantic calibration, semantic residual correction, and native visual evidence are progressively introduced to examine their respective contributions on CamCrack789 and CrackMap.}
\label{tab:4}
\end{table}

As shown in Table~\ref{tab:4}, optimizing the native semantic projection of SAM3 improves Crack IoU by 2.50 and 3.55 percentage points on the two datasets. The frozen prompt-conditioned representation therefore contains substantial crack information, while target-domain supervision further calibrates the native decision mapping acquired through general pretraining.

Adding semantic-only residual correction after semantic calibration further raises Crack IoU to 73.53\% on CamCrack789 and 63.30\% on CrackMap. The gain reaches 4.14 percentage points on CrackMap, showing that local nonlinear correction of the existing semantic state is itself effective.

Introducing native visual evidence further increases Crack IoU to 74.71\% and 65.78\%, gains of 1.18 and 2.48 percentage points over the semantic-only residual. On CrackMap, clDice also rises from 85.14\% to 89.09\%. The visual representation therefore provides additional conditional information for local correction of the calibrated semantic decision, consistent with the representation relationship revealed by the probing analysis in Section~\ref{sec:preliminary}.

\subsubsection{Residual Reconciliation versus Direct Feature Fusion}\label{sec:direct-fusion}

The preceding experiment shows that visual evidence produces gains in the residual branch. To separate the effects of feature quantity from decision structure, we construct a direct semantic--visual fusion baseline with a similar parameter count. After projecting the two representations independently, this baseline directly predicts a complete segmentation logit:
\begin{equation}
Z_{\mathrm{fusion}}=
h(F_p,F_v), \label{eq:25}
\end{equation}

CoRe instead always retains the existing semantic belief as its prediction basis, following Eq.~\eqref{eq:4}. Table~\ref{tab:5} reports average results on CamCrack789 and CrackMap.

\begin{table}[pos=t]
\centering
\scriptsize
\setlength{\tabcolsep}{3.5pt}
\renewcommand{\arraystretch}{1.08}
\resizebox{\textwidth}{!}{%
\begin{tabular}{lcccccc}
\toprule
\textbf{Method} & \textbf{Semantic prior retained} & \textbf{Visual evidence} & \textbf{Prediction form} & \textbf{Crack IoU} & \textbf{F1} & \textbf{clDice} \\
\midrule
SAM3 & $\checkmark$ &  & Native prediction & 63.05 & 74.95 & 85.74 \\
Semantic calibration & $\checkmark$ &  & Calibrated prediction & 66.07 & 81.60 & 89.37 \\
Direct S+V fusion &  & $\checkmark$ & Full prediction & 66.04 & 81.16 & 89.33 \\
S-only residual & $\checkmark$ &  & Residual correction & 67.96 & 81.88 & 88.77 \\
S+V residual & $\checkmark$ & $\checkmark$ & Residual correction & 68.55 & 81.53 & 90.52 \\
CoRe-SAM3 (ours) & $\checkmark$ & $\checkmark$ & Bounded residual correction & \textbf{70.24} & \textbf{82.10} & \textbf{91.89} \\
\bottomrule
\end{tabular}%
}
\caption{Comparison of strategies for incorporating semantic and visual representations. Direct semantic--visual fusion predicts a new segmentation output from the two representations, whereas residual variants preserve the calibrated semantic prediction and use additional representations for prediction correction. Results are averaged over CamCrack789 and CrackMap.}
\label{tab:5}
\end{table}

In Table~\ref{tab:5}, direct S+V fusion obtains 66.04\% Crack IoU, nearly identical to the 66.07\% of semantic calibration, with similar F1 and clDice. This result corroborates the cross-domain $[S,V]\rightarrow E$ probing in Section~\ref{sec:preliminary}: direct combination does not convert the native visual feature into a stronger target-domain prediction.

When the semantic prior is retained, the S-only residual increases Crack IoU to 67.96\%. Adding visual evidence further improves it to 68.55\% and raises clDice from 88.77\% to 90.52\%. The complete bounded reconciliation design ultimately reaches 70.24\% Crack IoU and 91.89\% clDice.

The adaptation benefit of $F_v$ therefore depends on its decision role. Direct fusion allows the visual and semantic representations to jointly reconstruct a complete prediction, whereas CoRe organizes the visual representation as conditional evidence under the existing semantic state. The results in Table~\ref{tab:5} support this asymmetric use of visual evidence.

\subsubsection{Effect of Conservative Residual Constraints}\label{sec:residual-constraints}

CoRe uses zero initialization, a bounded residual, and residual regularization to constrain the freedom of the new branch to deviate from the existing semantic belief. Table~\ref{tab:6} progressively introduces these constraints and reports Crack IoU, correction rate (CR), and damage rate (DR). For pixels misclassified by native SAM3, the correction rate is
\begin{equation}
\mathrm{CR}=N_{\mathrm{wrong}\rightarrow\mathrm{correct}} / N_{\mathrm{native\ wrong}}, \label{eq:26}
\end{equation}
where $N_{\mathrm{wrong}\rightarrow\mathrm{correct}}$ is the number of pixels that transition from an incorrect state to a correct state after CoRe. For pixels originally classified correctly by SAM3, the damage rate is
\begin{equation}
\mathrm{DR}=N_{\mathrm{correct}\rightarrow\mathrm{wrong}} / N_{\mathrm{native\ correct}}. \label{eq:27}
\end{equation}

\begin{table}[pos=t]
\centering
\begin{tabular}{lcccccc}
\toprule
\textbf{Variant} & \textbf{Zero initialization} & \textbf{Bounded residual} & $\mathcal L_{\mathrm{res}}$ & \textbf{Crack IoU} & \textbf{CR} $\downarrow$ & \textbf{DR} $\downarrow$ \\
\midrule
Unconstrained residual &  &  &  & 68.21 & 30.64\% & 0.33\% \\
+ Zero initialization & $\checkmark$ &  &  & 68.55 & 30.11\% & 0.29\% \\
+ Bounded residual & $\checkmark$ & $\checkmark$ &  & 69.16 & 29.40\% & 0.19\% \\
CoRe-SAM3 & $\checkmark$ & $\checkmark$ & $\checkmark$ & \textbf{70.24} & \textbf{29.16\%} & \textbf{0.18\%} \\
\bottomrule
\end{tabular}%
\caption{Effect of conservative constraints on residual reconciliation. Zero initialization, bounded residual prediction, and residual regularization are progressively introduced. CR and DR denote the correction rate on originally misclassified pixels and the damage rate on originally correctly classified pixels, respectively. Results are averaged over CamCrack789 and CrackMap.}
\label{tab:6}
\end{table}

Table~\ref{tab:6} shows that Crack IoU increases as the constraints become stronger. CR decreases slightly from 30.64\% to 29.16\%, while DR decreases from 0.33\% to 0.18\%. The performance improvement coincides with a lower damage rate, indicating greater selectivity in prediction changes.

Zero initialization lets the new branch start from the existing semantic solution. The bounded residual further limits the magnitude of each local correction and substantially reduces DR. Residual regularization suppresses inefficient deviations and achieves the highest Crack IoU. Higher performance therefore corresponds to more selective correction, consistent with the CoRe formulation of reconciliation as a conservative belief update.

\subsubsection{Effect of Crack-Aware Optimization}\label{sec:crack-aware-objective}

To separate the effect of representation reconciliation from that of the loss function, Table~\ref{tab:7} compares standard BCE plus Dice, the addition of outer-ring weighting, and the further addition of residual regularization.

\begin{table}[pos=t]
\centering
\begin{tabular}{lccccc}
\toprule
\textbf{Loss configuration} & \textbf{Crack IoU} & \textbf{Precision} & \textbf{Recall} & \textbf{PAR} & \textbf{clDice} \\
\midrule
BCE + Dice & 67.63 & 75.36 & 88.50 & 1.243 & 90.28 \\
+ Ring weighting & 69.16 & 79.09 & 85.32 & 1.150 & 91.39 \\
+ Residual regularization & \textbf{70.24} & \textbf{79.59} & \textbf{86.74} & \textbf{1.129} & \textbf{91.89} \\
\bottomrule
\end{tabular}%
\caption{Effect of crack-aware optimization objectives. The comparison progressively introduces crack-boundary ring weighting and residual regularization on top of standard BCE and Dice losses to evaluate their effects on foreground recovery, suppression of false activations, and structural preservation.}
\label{tab:7}
\end{table}

With standard BCE plus Dice, the model obtains high Recall, a Precision of 75.36\%, and a PAR of 1.243, indicating pronounced foreground expansion. Adding ring weighting slightly reduces Recall but increases Precision to 79.09\% and lowers PAR to 1.150, while substantially improving both Crack IoU and clDice. Stronger supervision on background pixels near cracks therefore suppresses boundary expansion and false activations on nearby crack-like patterns.

Residual regularization preserves Precision and recovers part of the lost Recall, ultimately reaching 70.24\% Crack IoU and 91.89\% clDice. Crack-aware optimization thus constrains the decision behavior of the reconciliation branch around crack boundaries and nearby background regions, yielding a better balance between recovering missing foreground and suppressing erroneous expansion.

\subsection{Decision Correction Analysis}\label{sec:correction-analysis}

The preceding results show that CoRe substantially improves final segmentation performance. We further identify which pixels the model modifies and in which direction. The analysis characterizes constrained changes to the existing semantic belief through both discrete prediction-state transitions and continuous response changes.

\subsubsection{Correction of Native Errors and Preservation of Correct Decisions}\label{sec:prediction-transitions}

\begin{figure}[pos=t]
\centering
\includegraphics[width=0.98\textwidth]{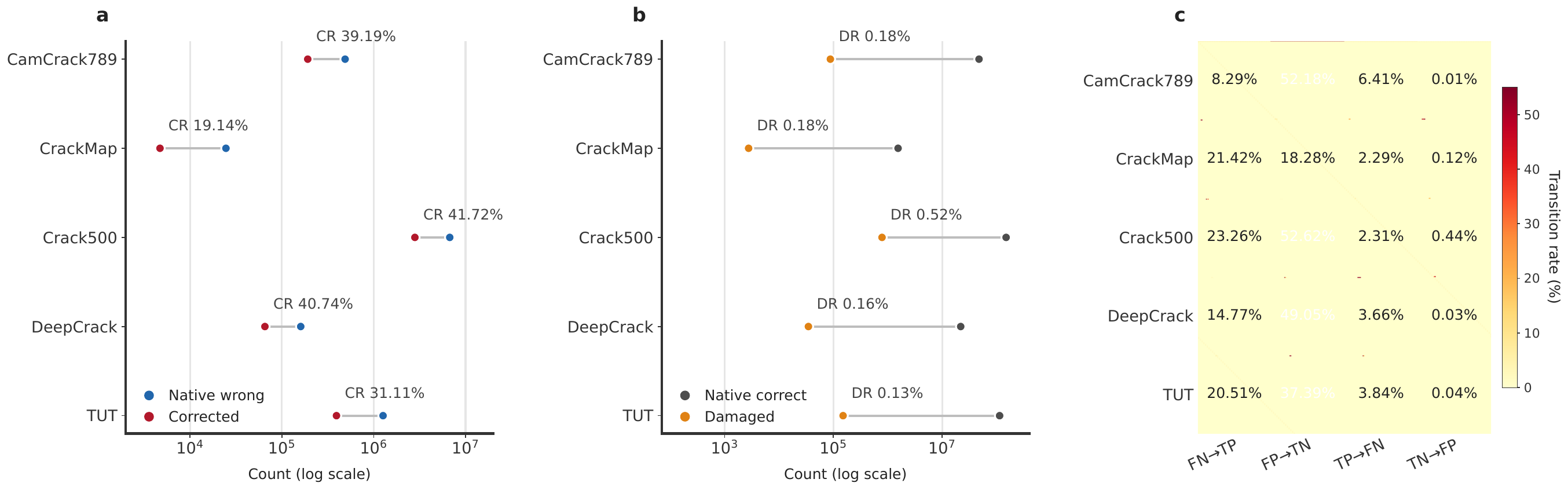}
\caption{Pixel-level prediction transitions from native SAM3 to CoRe-SAM3 across five crack datasets. (a) Log-scale paired counts of pixels misclassified by native SAM3 and those corrected by CoRe-SAM3, together with the corresponding correction rate (CR). (b) Log-scale paired counts of pixels correctly classified by native SAM3 and those damaged by CoRe-SAM3, together with the corresponding damage rate (DR). (c) Class-conditional transition rates for false-negative-to-true-positive (FN$\rightarrow$TP), false-positive-to-true-negative (FP$\rightarrow$TN), true-positive-to-false-negative (TP$\rightarrow$FN), and true-negative-to-false-positive (TN$\rightarrow$FP) changes.}
\label{fig:4}
\end{figure}

Fig.~\ref{fig:4}(a) shows that CoRe corrects a substantial fraction of native errors in all five data domains. The CR values on CamCrack789, CrackMap, Crack500, DeepCrack, and TUT are 39.19\%, 19.14\%, 41.72\%, 40.74\%, and 31.11\%, respectively, with a five-domain mean of 34.38\%. The corresponding DR values in Fig.~\ref{fig:4}(b) are 0.18\%, 0.18\%, 0.52\%, 0.16\%, and 0.13\%, with a mean of 0.234\%. These results reveal a pronounced asymmetry between correction and damage.

Fig.~\ref{fig:4}(c) further separates transitions involving FN and FP errors. On CamCrack789, Crack500, and DeepCrack, CoRe converts 52.18\%, 52.62\%, and 49.05\% of native FP pixels into TN pixels, respectively, demonstrating strong suppression of crack-like backgrounds. FN$\rightarrow$TP transitions reach 21.42\%, 23.26\%, and 20.51\% on CrackMap, Crack500, and TUT, respectively, showing that the model also recovers crack regions that native SAM3 insufficiently activates.

The substantial class imbalance in crack segmentation makes overall DR largely depend on the dominant TN pixels, so Fig.~\ref{fig:4}(c) also reports class-conditional transition rates. TN$\rightarrow$FP remains below 0.44\% on every dataset, while TP$\rightarrow$FN ranges from 2.29\% to 6.41\%. CoRe therefore preserves correctly predicted backgrounds at a very high rate while suppressing a small fraction of existing crack TP pixels. Overall damage rate and class-conditional transitions jointly characterize the decision-preservation capability of conservative adaptation.

\subsubsection{Direction of Prediction Adjustment}\label{sec:adjustment-direction}

To analyze how CoRe changes the continuous SAM3 response, we define the total logit adjustment from native SAM3 to the final CoRe output as
\begin{equation}
A(x) = Z_{\mathrm{final}}(x)-Z_{\mathrm{sam}}(x) = \underbrace{Z_{\mathrm{cal}}(x)-Z_{\mathrm{sam}}(x)}_{\text{semantic calibration}} + \underbrace{\Delta Z_{\mathrm{reco}}(x)}_{\text{conditional reconciliation}} \label{eq:28}
\end{equation}
where $A$ denotes the total prediction change produced by the complete adaptation process and includes both semantic calibration and conditional reconciliation. In contrast, $\Delta Z_{\mathrm{rec}}$ in Section~\ref{sec:bounded-reconciliation} denotes only the reconciliation residual constrained to $[-s,s]$. We further define
\begin{equation}
R^{-}=P(\Delta Z<0),  \qquad R^{+} = P(\Delta Z>0) \label{eq:29}
\end{equation}
as the proportions of negative and positive adjustments, respectively, within a pixel region. The error-related directions are highly consistent across data domains. Negative adjustments dominate FP regions of native SAM3, whereas positive adjustments dominate FN regions.

Table~\ref{tab:8} summarizes the directional proportions of total logit adjustments in FP and FN regions of native SAM3.

\begin{table}[pos=t]
\centering
\begin{tabular}{cccccc}
\toprule
\textbf{Region / Statistic} & \textbf{CamCrack789} & \textbf{CrackMap} & \textbf{Crack500} & \textbf{DeepCrack} & \textbf{TUT} \\
\midrule
FP $R^{-}$ & 98.72\% & 94.45\% & 88.64\% & 97.78\% & 96.54\% \\
FN $R^{+}$ & 68.95\% & 90.85\% & 91.49\% & 74.84\% & 87.46\% \\
\bottomrule
\end{tabular}%
\caption{Directional statistics of prediction adjustments in false-positive and false-negative regions. $R^{-}$ denotes the proportion of false-positive pixels receiving negative logit adjustments, while $R^{+}$ denotes the proportion of false-negative pixels receiving positive logit adjustments. FP and FN regions are determined from native SAM3 predictions.}
\label{tab:8}
\end{table}

As shown in Table~\ref{tab:8}, negative adjustments are applied to 88.64\%--98.72\% of FP pixels across the five domains, while positive adjustments are applied to 68.95\%--91.49\% of FN pixels. These directions match the required error corrections:
\begin{equation}
\mathrm{FP} \rightarrow \Delta Z<0, \quad \mathrm{FN} \rightarrow \Delta Z>0. \label{eq:30}
\end{equation}

CoRe therefore uses the current semantic state and local visual evidence to apply negative adjustments to FP errors and positive adjustments to FN errors.

\begin{figure}[pos=t]
\centering
\includegraphics[width=0.98\textwidth]{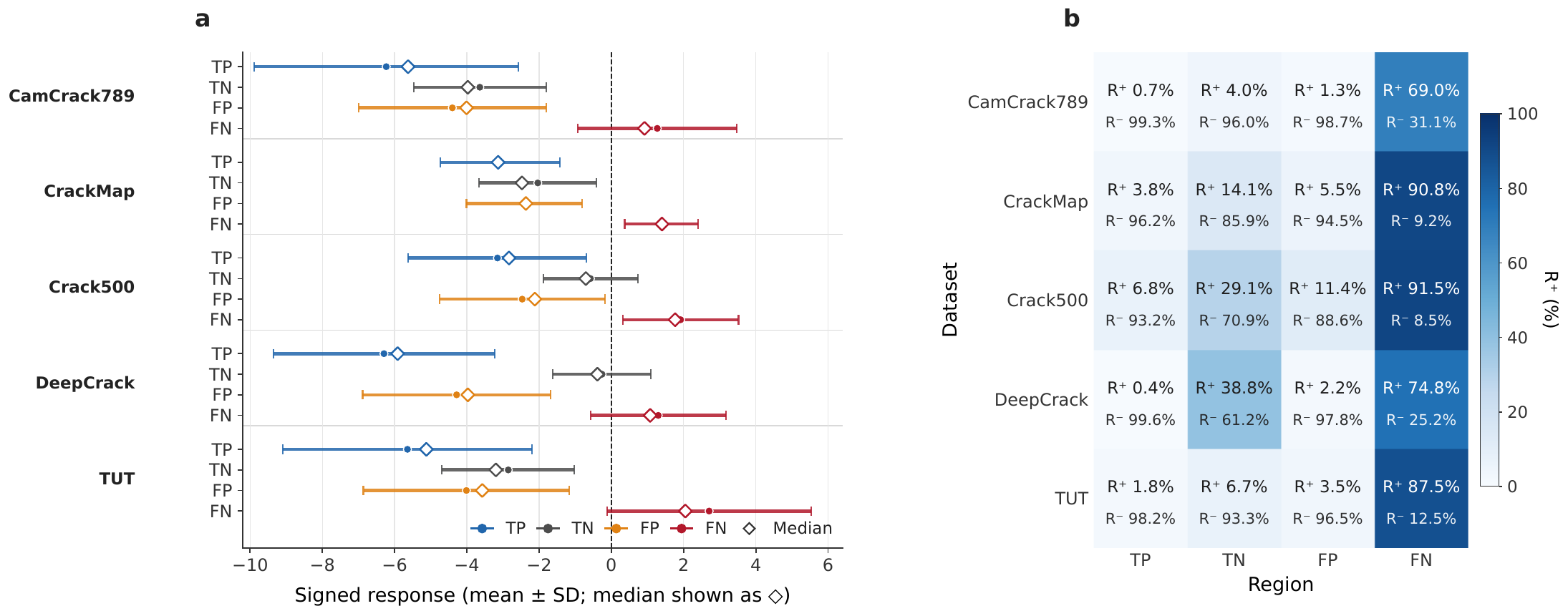}
\caption{Directional analysis of prediction adjustments induced by CoRe-SAM3. The total prediction adjustment is defined as $A=Z_{\mathrm{final}}-Z_{\mathrm{sam}}$ and incorporates both semantic calibration and conditional reconciliation. (a) Mean $\pm$ standard deviation and median signed adjustments for true-positive (TP), true-negative (TN), false-positive (FP), and false-negative (FN) regions under the native SAM3 prediction. (b) Dataset-wise proportions of positive and negative adjustments, denoted by $R^{+}$ and $R^{-}$, for the same four regions.}
\label{fig:5}
\end{figure}

Fig.~\ref{fig:5}(a) reports the mean, standard deviation, and median of total adjustments in the four regions. FP pixels predominantly receive negative adjustments, whereas FN pixels predominantly receive positive adjustments. Fig.~\ref{fig:5}(b) further reports per-domain $R^{+}$ and $R^{-}$. Only approximately 0.4\%--6.8\% of TP pixels receive positive adjustments, reflecting the recalibration of the overall target-domain response distribution by semantic calibration. Together with the TP$\rightarrow$FN transition rates of 2.29\%--6.41\% in Section~\ref{sec:prediction-transitions}, these results show that most TP adjustments change prediction confidence while preserving the final class state.

Overall, CoRe behaves as a combination of target-domain response recalibration and conditional local compensation. Semantic calibration first changes the global decision distribution of the general SAM3 response in the target crack domain. Semantic--visual reconciliation then further adjusts local regions that require recovery or suppression. This behavior agrees with the method design that separates global semantic decision retargeting from local conditional reconciliation.

\subsection{Hyperparameter Sensitivity Analysis}\label{sec:sensitivity}

We evaluate the sensitivity of CoRe-SAM3 to key hyperparameters on CamCrack789 and CrackMap using a one-factor-at-a-time (OFAT) protocol. Each run changes only one parameter while fixing all others at the default configuration, namely $s=0.5$, $r=3$, $\omega=4$, $\lambda_d=0.25$, and $\lambda_r=0.05$. The candidate sets are $\{0.1,0.25,0.5,0.75,1.0\}$ for the residual magnitude limit $s$, $\{1,2,3,4,5\}$ for the morphological radius $r$, $\{1,2,4,6,8\}$ for the additional outer-ring weight $\omega$, $\{0,0.1,0.25,0.5,1.0\}$ for the Dice loss weight $\lambda_d$, and $\{0,0.01,0.05,0.1,0.2\}$ for the residual regularization weight $\lambda_r$. All experiments use test-set mIoU as the evaluation metric.

\begin{figure}[pos=t]
\centering
\includegraphics[width=0.5\textwidth]{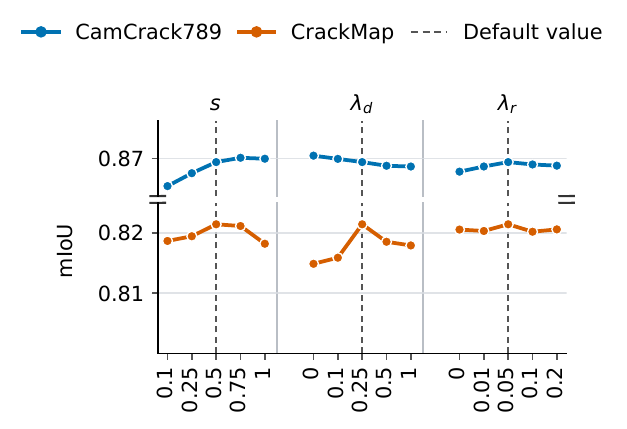}
\caption{One-factor-at-a-time sensitivity of adapted test mIoU to the residual scale $s$, Dice loss weight $\lambda_d$, and residual regularization weight $\lambda_r$ on CamCrack789 and CrackMap. Dashed vertical lines indicate the default settings, while all remaining hyperparameters are fixed at their default values. The broken y-axis omits the interval $0.83\text{--}0.86$ for improved visibility.}
\label{fig:6}
\end{figure}

Fig.~\ref{fig:6} analyzes the effects of the residual magnitude limit $s$, Dice loss weight $\lambda_d$, and residual regularization weight $\lambda_r$. None of these parameters causes a substantial performance change on CamCrack789 within the evaluated ranges. The default configuration obtains an mIoU of 0.8695. The settings $s=0.75$ and $\lambda_d=0$ obtain 0.8701 and 0.8704, respectively, each differing from the default by less than 0.001. For $\lambda_r$, the default value of 0.05 yields the highest observed result among the candidates. CrackMap exhibits a similar trend. The default $s=0.5$ obtains an mIoU of 0.8214, and varying $\lambda_d$ or $\lambda_r$ produces similarly small changes. These fixed-seed observations show that CoRe-SAM3 maintains stable performance over broad ranges of the residual magnitude and the two loss weights.

\begin{figure}[pos=t]
\centering
\includegraphics[width=0.5\textwidth]{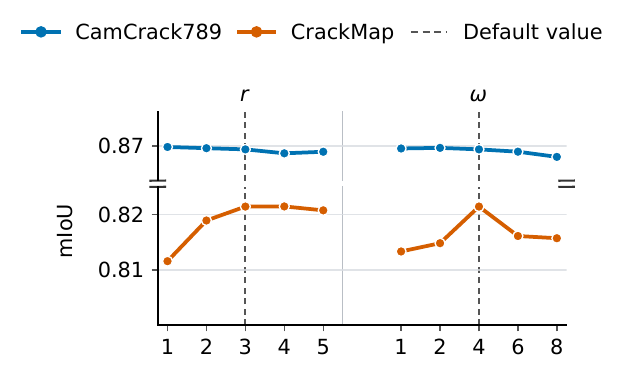}
\caption{One-factor-at-a-time sensitivity of adapted test mIoU to the morphological radius $r$ and outside-ring weight $\omega$ on CamCrack789 and CrackMap. Dashed vertical lines indicate the default settings, with all other hyperparameters fixed. The broken y-axis omits the interval $0.83\text{--}0.86$ for improved visibility.}
\label{fig:7}
\end{figure}

Fig.~\ref{fig:7} further examines the morphological radius $r$ and additional weight $\omega$ of the outer weighted region. On CamCrack789, $r=1$ yields the highest observed mIoU of 0.8699, compared with 0.8695 for the default $r=3$. For $\omega$, the setting $\omega=2$ and the default $\omega=4$ yield 0.8697 and 0.8695, respectively. The differences among settings are small. CrackMap is similarly insensitive to both parameters. The default configuration obtains 0.8214 mIoU, the highest observed value is 0.82146 at $r=4$, and $\omega=4$ yields the highest result in the current candidate set. The spatial extent and additional weight of outer-region supervision therefore have little effect on performance within reasonable ranges.

Taken together, Figs.~\ref{fig:6} and~\ref{fig:7} show limited performance variation across the five key hyperparameters on both data domains. CoRe-SAM3 therefore obtains stable results without a narrow parameter search. We use $s=0.5$, $r=3$, $\omega=4$, $\lambda_d=0.25$, and $\lambda_r=0.05$ as the unified default configuration to maintain consistent settings and stable performance across domains.

\subsection{Qualitative Analysis}\label{sec:qualitative}

\begin{figure}[pos=t]
\centering
\includegraphics[width=0.98\textwidth]{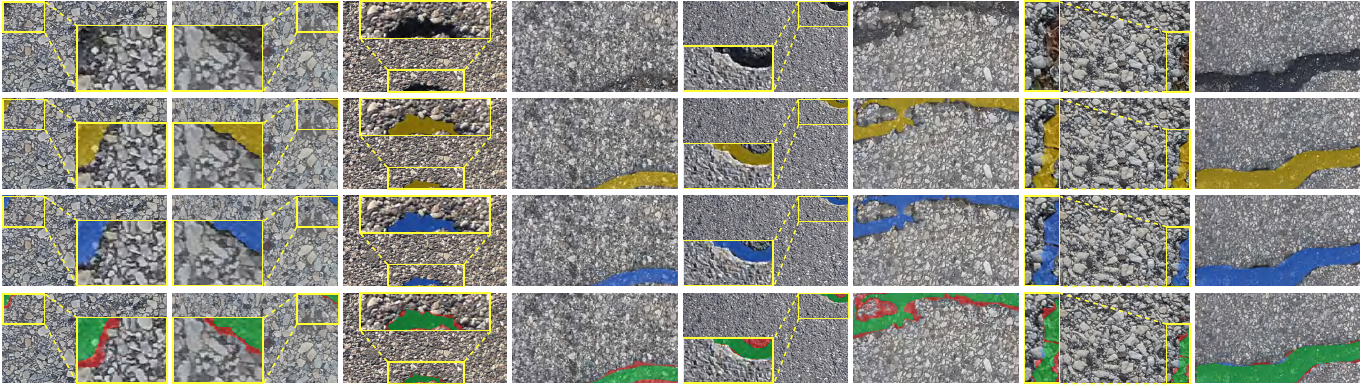}
\caption{Qualitative examples of recovering weak and incomplete crack responses. Compared with native SAM3, CoRe-SAM3 restores crack regions with weak semantic responses and improves the continuity of thin or locally interrupted structures while largely preserving the original crack topology.}
\label{fig:8}
\end{figure}

\begin{figure}[pos=t]
\centering
\includegraphics[width=0.98\textwidth]{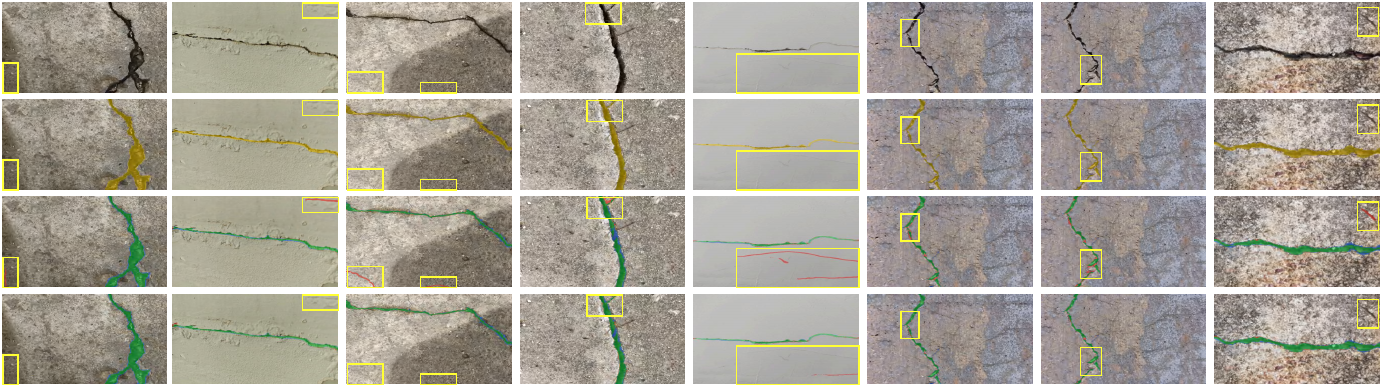}
\caption{Qualitative examples of suppressing crack-like false-positive responses. CoRe-SAM3 conditionally reconciles the existing semantic prediction with native visual evidence, reducing erroneous activations caused by joints, texture boundaries, shadows, and other crack-like background structures.}
\label{fig:9}
\end{figure}

\begin{figure}[pos=t]
\centering
\includegraphics[width=0.98\textwidth]{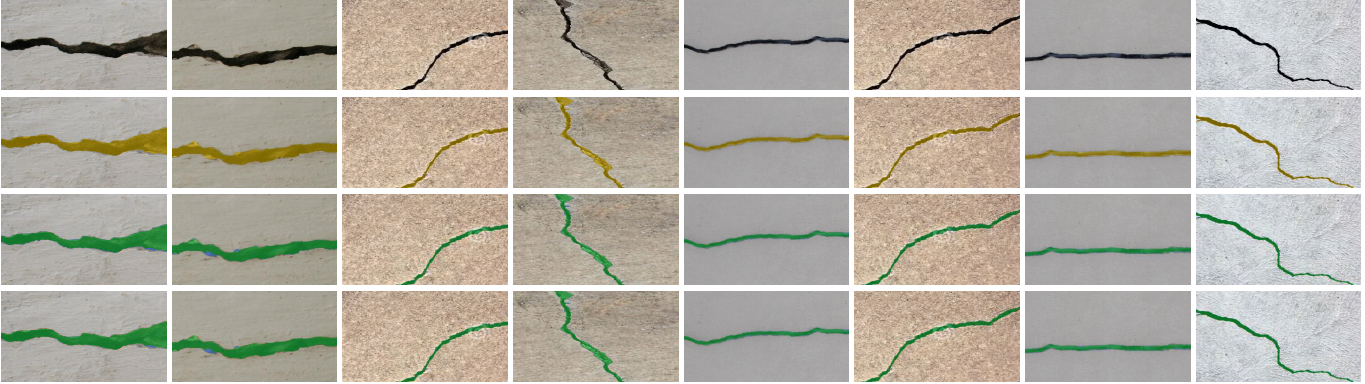}
\caption{Qualitative examples of preserving reliable native SAM3 predictions during target-domain adaptation. For regions already segmented correctly by native SAM3, CoRe-SAM3 applies localized response and boundary adjustments while preserving the overall prediction structure, illustrating the conservative behavior of the proposed reconciliation strategy.}
\label{fig:10}
\end{figure}

Figs.~\ref{fig:8}--\ref{fig:10} qualitatively compare three representative effects of CoRe-SAM3 on native SAM3 predictions: recovery of weak responses and incomplete cracks, suppression of false activations on crack-like backgrounds, and preservation of correct native predictions. Each figure presents different samples by column. From top to bottom, the four rows show the input image, ground truth, native SAM3 prediction, and CoRe-SAM3 prediction. Yellow boxes indicate representative local regions. Green, red, and blue in the prediction overlays denote TP, FP, and FN pixels, respectively. Beyond comparing final segmentation contours, these examples examine how CoRe behaves under different native prediction states and whether this behavior agrees with conditional reconciliation.

Fig.~\ref{fig:8} shows typical cases in which native SAM3 responds insufficiently to low-contrast, thin, or locally discontinuous cracks. SAM3 usually identifies the main crack direction and body, indicating that its prompt-conditioned semantic representation already contains strong crack-semantic information. Its response can still decay or disappear where crack width decreases, local contrast weakens, or small branches emerge. After CoRe adaptation, these missing regions are recovered, some disconnected crack segments become continuous, and the location and overall direction of the main crack remain largely unchanged. In the thin branches and local interruptions highlighted by yellow boxes, blue FN regions decrease substantially, while new responses extend mainly along existing crack structures without unconstrained expansion into the surrounding background. CoRe therefore does not generate a segmentation unrelated to the original result. It uses local visual evidence to compensate for insufficient responses under the existing semantic state. This behavior agrees with the improvements in Recall and clDice: Recall reflects recovery of missed crack regions, while clDice shows that the added predictions improve connectivity and topological completeness.

Fig.~\ref{fig:9} presents false activations caused by crack-like backgrounds. Joints, texture boundaries, shadow edges, and local linear structures can resemble cracks, so native SAM3 may over-respond in these regions despite its strong semantic judgment of cracks. In several examples, SAM3 correctly covers the true crack body but also predicts neighboring joints, regular textures, or high-contrast edges. CoRe-SAM3 preserves true crack responses while selectively weakening activations that lack sufficient local visual support, substantially reducing red FP regions. This suppression does not arise from globally reducing prediction strength because true crack regions do not undergo similar large-scale contraction. Changes concentrate on locally unreliable positions in the original prediction. This behavior agrees with the correction statistics: 88.64\%--98.72\% of adjustments in FP regions are negative and produce high FP$\rightarrow$TN transition rates. Native visual evidence in CoRe therefore does not act as a second independent segmentation that overrides semantic prediction. It evaluates whether an existing activation receives sufficient local image support under the current semantic state and applies a suppressive correction accordingly.

Fig.~\ref{fig:10} examines whether target-domain adaptation damages correct native predictions. The native SAM3 results in these examples already agree closely with the ground truth and correctly recover the crack body, extension direction, and main topological relationships. CoRe-SAM3 does not substantially reconstruct these predictions. Its changes mainly involve local boundary contraction, width adjustment, and small response corrections. Comparing the third and fourth rows shows that most green TP regions are preserved and the overall crack geometry changes little. The residual branch therefore does not continually seek large prediction changes and maintains a small correction magnitude when the current semantic prediction is reliable. This observation agrees with the bounded residual, residual regularization, and low damage rate, showing that constrained correction improves target-domain adaptation while limiting damage to correct SAM3 decisions.

Figs.~\ref{fig:8}--\ref{fig:10} jointly reveal three complementary behaviors. CoRe recovers missing responses and improves structural continuity for true crack regions with insufficient semantic responses. It applies negative corrections to suppress false activations on crack-like backgrounds. It preserves the overall structure of reliable native predictions and makes only limited local adjustments. These behaviors correspond to FN recovery, FP suppression, and TP preservation, respectively. The gain of CoRe therefore does not arise from globally reconstructing SAM3 outputs but from selectively correcting the existing semantic belief. The qualitative results further support the central design: semantic evidence retains the primary prediction role, while native visual evidence provides a conditional basis for local correction under the current semantic state, enabling conservative and effective semantic--visual reconciliation.

\subsection{Efficiency Analysis}\label{sec:efficiency}

CoRe reuses semantic and visual representations already computed inside SAM3 and performs target-domain adaptation with a lightweight reconciliation branch. Table~\ref{tab:9} reports end-to-end computational cost without feature caching.

\begin{table}[pos=t]
\centering
\begin{tabular}{lccccc}
\toprule
\textbf{Method} & \textbf{Trainable Params} & \textbf{Params} & \textbf{FLOPs} & \textbf{Latency} & \textbf{Peak GPU Memory} \\
\midrule
SAM3 & 0 & 840.510 M & 3.672 T & 451.2 ms & 3889.7 MiB \\
CoRe-SAM3 & 18.914 K & 840.529 M & 3.678 T & 539.1 ms & 3889.7 MiB \\
\bottomrule
\end{tabular}%
\caption{End-to-end computational efficiency of SAM3 and CoRe-SAM3. Latency and peak GPU memory are measured during complete model inference without cached SAM3 features. FLOPs are estimated over the complete forward path. Trainable parameters include only the parameters updated during task adaptation.}
\label{tab:9}
\end{table}

The efficiency evaluation runs the complete SAM3 forward pass without reading any precomputed feature cache. Measurements use an NVIDIA RTX A2000 GPU with 12 GB memory and bfloat16 autocast. A randomly selected image serves as input. We perform three warm-up iterations, time ten inference runs, and report the mean. FLOPs are estimated over the complete forward path with the PyTorch profiler.

As shown in Table~\ref{tab:9}, CoRe adds 18.914 K trainable parameters, approximately 0.0023\% of the total SAM3 parameter count. Total FLOPs increase from 3.672 T to 3.678 T, corresponding to an additional 6.211 GFLOPs or approximately 0.16\%, while peak memory remains unchanged at 3889.7 MiB. In terms of model capacity and theoretical computation, the new CoRe branch is therefore lightweight relative to the SAM3 backbone.

End-to-end latency increases from 451.2 ms to 539.1 ms, or approximately 19.5\%. This increase is substantially larger than the relative FLOP change, indicating that low-dimensional projections, feature access, and multiple small operators incur practical overhead beyond their theoretical FLOP share. The efficiency of CoRe lies primarily in reusing a single visual backbone, introducing very few parameters, and maintaining the same peak memory. The current implementation incurs an approximately 19.5\% increase in practical inference latency.

Overall, the very small number of additional parameters, approximately 0.16\% increase in theoretical computation, and unchanged peak memory show that the performance gain of CoRe primarily results from reorganizing existing internal evidence in SAM3.

\subsection{Discussion}\label{sec:discussion}

\subsubsection{Implications of Conditional Reconciliation}\label{sec:implications}

For a vision foundation model that already contains a strong task-semantic prior, effective downstream adaptation depends on organizing the functional relationships among existing internal representations, not solely on increasing representation capacity. Five-domain probing shows that the prompt-conditioned semantic representation of SAM3 performs most crack discrimination, while the native visual representation is better suited to serving as supplementary evidence conditioned on the current semantic state. Consistent with this diagnosis, direct semantic--visual fusion does not provide a stable advantage, whereas residual reconciliation that retains the semantic prior produces consistent gains. Effective use of internal representations in foundation models therefore depends on both the information they contain and the decision roles through which they affect final predictions.

CoRe instantiates this functional distinction as a constrained prediction update. Semantic calibration first adjusts the target-domain decision mapping of the existing semantic representation, and conditional reconciliation applies only a limited correction to the calibrated prediction. Progressively introducing zero initialization, a bounded residual, and residual regularization improves Crack IoU while reducing damage to correct native predictions. Five-domain prediction transitions and response adjustments further show that CoRe predominantly applies negative adjustments to FP regions and positive adjustments to FN regions while preserving most existing correct decisions. Its performance gain therefore arises from selective correction of the existing semantic belief instead of relearning a complete segmentation decision.

These results support an adaptation principle for models with strong pretrained priors: first analyze the relative functions of internal representations, then constrain the decision authority of new representations and adaptation branches accordingly. CoRe implements this principle through semantic calibration and conditional semantic--visual reconciliation. With only 18.914 K trainable parameters, it improves the five-domain average Crack IoU from 62.34\% to 70.47\%.

\subsubsection{Limitations and Future Work}\label{sec:limitations}

This study focuses on adapting SAM3 to crack segmentation and analyzes the functional relationship between the prompt-conditioned semantic representation and a native visual representation at the same spatial scale. The five datasets cover diverse imaging conditions, background textures, and crack morphologies, but the current conclusions are grounded primarily in curvilinear crack segmentation and the internal representation system of SAM3. Future work can evaluate representation-role organization and conditional reconciliation on other segmentation tasks with fine spatial structures and on other vision foundation models. It can also investigate multilevel native visual representations, dynamic residual control, and more efficient access to internal features to extend the generality and implementation efficiency of the framework.

\section{Conclusion}\label{sec:conclusion}

This paper studies target-domain adaptation of SAM3 for crack segmentation by analyzing the roles of prompt-conditioned semantic evidence and native visual evidence within the model. Five-domain probing shows that the semantic representation strongly carries crack-task information, whereas the native visual representation is weaker for independent prediction and lacks a stable error-discrimination gain when directly combined with the semantic representation. Based on this diagnosis, we propose CoRe, which decomposes adaptation into target-domain semantic calibration and conditional semantic--visual reconciliation. The former retains and recalibrates the existing SAM3 semantic decision, while the latter uses spatially aligned native visual evidence to apply a zero-initialized, bounded, and regularized residual correction to prediction errors.

Across five crack datasets, CoRe-SAM3 improves the average Crack IoU from 62.34\% to 70.47\% and clDice from 81.98\% to 89.24\%, while adding 18.914 K trainable parameters. Ablation results show that direct semantic--visual fusion performs similarly to semantic calibration, whereas residual reconciliation consistently benefits from retaining the semantic prior. As zero initialization, a bounded residual, and residual regularization are progressively introduced, performance increases while the damage rate decreases. Across five domains, CoRe corrects an average of 34.38\% of native errors with an average damage rate of approximately 0.23\% on correctly classified native pixels. Continuous-response analysis further shows that FP regions predominantly receive negative adjustments, while FN regions predominantly receive positive adjustments.

Together, these results show that adaptation of a vision foundation model with a strong task-semantic prior depends on the relative capabilities of its internal evidence and the decision authority assigned to each representation. CoRe organizes native visual evidence as a constrained correction conditioned on the semantic belief, providing an adaptation strategy for specialized fine-grained segmentation that preserves strong priors while selectively correcting their errors.

\section*{Acknowledgment}
This work would not have been possible without financial support from the the Science and Technology Innovation Team of Shaanxi Innovation Capability Support Plan (2020TD-005), the Xi'an Scientists $\&$ Engineers Workforce Building Project (2024JH-KGDW-0112), and the Key Project of the Natural Science Basic Research Program of Shaanxi Province (2025SYS-SYSZD-049). The authors are also grateful to the editors and anonymous reviewers for their sound and valuable suggestions on the manuscript.

\section*{Declaration of generative AI and AI-assisted technologies in the manuscript preparation process}
During the preparation of this work the authors used ChatGPT (OpenAI) for translation and language polishing. After using this tool/service, the authors reviewed and edited the content as needed and take full responsibility for the content of the manuscript.

\bibliographystyle{unsrt}
\bibliography{references}

\end{document}